\documentclass[10pt,twocolumn,letterpaper]{article}

\usepackage[pagenumbers]{cvpr} 

\usepackage{graphicx}
\usepackage{amsmath}
\usepackage{amssymb}
\usepackage{booktabs}

\usepackage[pagebackref,breaklinks,colorlinks]{hyperref}

\usepackage[capitalize]{cleveref}
\crefname{section}{Sec.}{Secs.}
\Crefname{section}{Section}{Sections}
\Crefname{table}{Table}{Tables}
\crefname{table}{Tab.}{Tabs.}
\Crefname{figure}{Figure}{Figures}
\crefname{figure}{Fig.}{Figs.}

\def\cvprPaperID{*****} 
\def\confName{CVPR}
\def\confYear{2022}

\begin{document}

\title{An Ensemble Approach for Facial Expression Analysis in Video}

\author{
Hong-Hai Nguyen$^1$, Van-Thong Huynh$^1$, Soo-Hyung Kim\thanks{Corresponding author}\\
Department of Artificial Intellegence Convergence\\
Chonnam National University\\
Gwangju, South Korea \\
{\tt\small honghaik14@gmail.com, \{vthuynh,shkim\}@jnu.ac.kr}
}
\maketitle
\def\thefootnote{1}\footnotetext{Equal contribution}

\begin{abstract}
    Human emotions recognization contributes to the development of human-computer interaction. The machines understanding human emotions in the real world will significantly contribute to life in the future. This paper will introduce the Affective Behavior Analysis in-the-wild (ABAW3) 2022 challenge. The paper focuses on solving the problem of the valence-arousal estimation and action unit detection. For valence-arousal estimation, we conducted two stages: creating new features from multimodel and temporal learning to predict valence-arousal. First, we make new features; the Gated Recurrent Unit (GRU) and Transformer are combined using a Regular Networks (RegNet) feature, which is extracted from the image. The next step is the  GRU combined with Local Attention to predict valence-arousal. The Concordance Correlation Coefficient (CCC) was used to evaluate the model.
\end{abstract}

\section{Introduction}
\label{sec:intro}

People's emotions affect their lives and work. Researchers are trying to create machines that can detect and analyze human emotions that contribute to the development of intelligent machines. Therefore, there are a lot of applications in life, such as medicine, health, tracking or driver fatigue, etc. ~\cite{kollias2022abaw}.

The practical applications have many challenges from an uncontrolled environment. However, data sources are increasingly various from social networks and applications. Besides, the deep learning network improved the analysis and recognition process. Therefore, the ABAW3 2022 ~\cite{kollias2022abaw} was organized for affective behavior analysis in the wild. The challenge includes four tasks: Valence-Arousal (VA) Estimation, Expression (Expr) Classification, Action Unit (AU) Detection, and Multi-Task-Learning (MTL). This paper only focuses on the VA task. In this task, participants will predict a valence-arousal dimension based on data from the video.

In this study, we utilize feature extraction from the deep learning model. The images are extracted feature from the RegNet network ~\cite{radosavovic2020designing}. Multimodel applied with the RegNet feature, which is the combination of Gated Recurrent Units (GRUs) ~\cite{cho2014properties}  and Transformer ~\cite{devlin2018bert} to create new features as stage 1. For stage 2, We utilized new features. The GRU was applied to get temporal features. Besides, local attention was used to improve the model. 

In this work, we focus on valence-arousal prediction. Our contributions in this paper are summarized as:
\begin{itemize}
    \item Utilization features from the deep learning model. 
    \item Using multimodal to create new features which increase speed training.
    \item Combination of Local Attention with GRU for sentiment analysis.
    \item Conducting experiments on different models to compare with baseline method.
\end{itemize}

The next parts of our paper are presented in the following sections: \autoref{sec:related_work} is related work, \autoref{sec:methodology} is methodology and the experimental results in \autoref{sec:experiments} and finally, the conclusion in \autoref{sec:conclustion}.

\section{Related work}
\label{sec:related_work}
In this section, we shortly summarize some datasets and works related to the problem of affective behavior in the previous challenge.
\subsection{Affect Annotation Dataset}
In the previous challenge \cite{kollias2021analysing, kollias2020analysing, kollias2021distribution, kollias2021affect, kollias2019expression, kollias2019face, kollias2019deep, zafeiriou2017aff}, the ABAW3 provides a large-scale dataset Aff-Wild2 for affective behavior analysis in-the-wild. The dataset used is the Aff-wild2 which was extended from Aff-wild1 \cite{zafeiriou2017aff}. The dataset contains annotations for challenges: Valence-Arousal regression, basic emotions, and Action Unit. Aff-wild2 expand the number of videos with 567 videos annotated by valence-arousal, 548 videos annotated by 8 expression categories, 547 videos annotated by 12 AUs, and 172,360 images are used that contain annotations of valence-arousal; 6 basic expressions, plus the neutral state, plus the ’other’ category; 12 action units.

\subsection{Affective Behavior Analysis in the wild}
The affective behavior analysis in the wild challenge has attracted a lot of researchers. Deng et al. \cite{deng2021iterative} applied deep ensemble models learned by a multi-generational self-distillation algorithm to improve emotion uncertainty estimation. About architectures, the author used features extractors from the efficient CNN model and applied GRU as temporal models. Wei Zhang et al. \cite{zhang2021prior} introduced multi-task recognition which is a streaming network by exploiting the hierarchical relationships between different emotion representations in the second ABAW challenge.  In the paper, Vu et al. \cite{vu2021multitask} used a multi-task deep learning model for valence-arousal estimation and facial expressions prediction. The authors applied the distillation knowledge architecture to train two networks: teacher and student model, because the dataset does not include labels for all the two tasks.  Another author, Kuhnke et al. \cite{kuhnke2020two} introduced a two-stream network for multi-task training. The model used the multimodel information extracted from audio and vision. The author \cite{deng2020multitask} solved two challenges of the competition. First, the problem is highly imbalanced in the dataset. Second, the datasets do not include labels for all three tasks. The author applied balancing techniques and proposed a Teacher-Student structure to learn from the imbalance labels to tackle the challenges.

\section{Methodology}
\label{sec:methodology}
For this section,  we introduce the proposed method for continuous emotion estimation. Our approach contains two stages: Create new features to increase training speed in \autoref{fig:va_unimodal} and use GRU to learn temporal information, illustrated in \autoref{fig:gru_att}.
Besides, Local Attention was applied to improve the model.

\subsection{Visual feature extraction}
Our visual feature is based on RegNet~\cite{radosavovic2020designing} architecture, a lightweight and efficient network. RegNet consists four stages to operate progressively reduced resolution with sequence of identical blocks. The pretrained weight from ImageNet~\cite{deng2009imagenet} was used as initial training, and the last three stages are unfreezing to learn new representation from facial data.

\subsection{Valence and arousal estimation}
\begin{figure}
  \centerline{\includegraphics[width=0.5\textwidth]{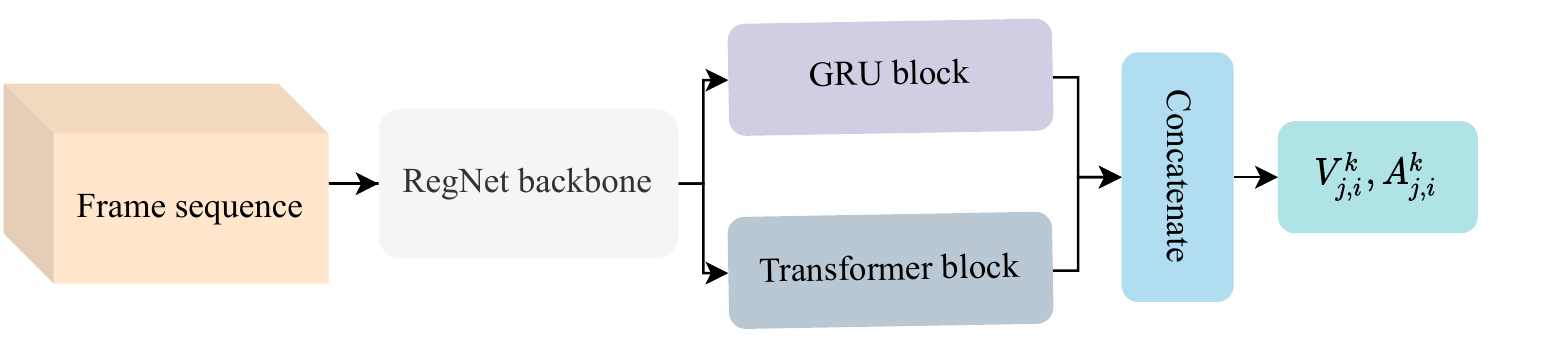}}
  \caption{Feature extraction architecture: Multimodal fusion with combined loss.}
  \label{fig:va_unimodal}
\end{figure}

Our temporal learning for unimodal consists of a combination of Gated recurrent units (GRU) block~\cite{cho2014properties} - a standard recurrent network, and Transformer block~\cite{vaswani2017attention} - attention based for sequential learning, as shown in~\autoref{fig:va_unimodal}. The representations from GRU and Transformer are concatenated to form a new feature vector and fed to a fully connected (FC) layer for producing valence and arousal scores. FC layers are also attached to GRU and Transformer blocks to obtain emotion scores for calculating loss function and combine to the final loss to optimize the whole system.

\begin{figure}
  \centerline{\includegraphics[width=0.5\textwidth]{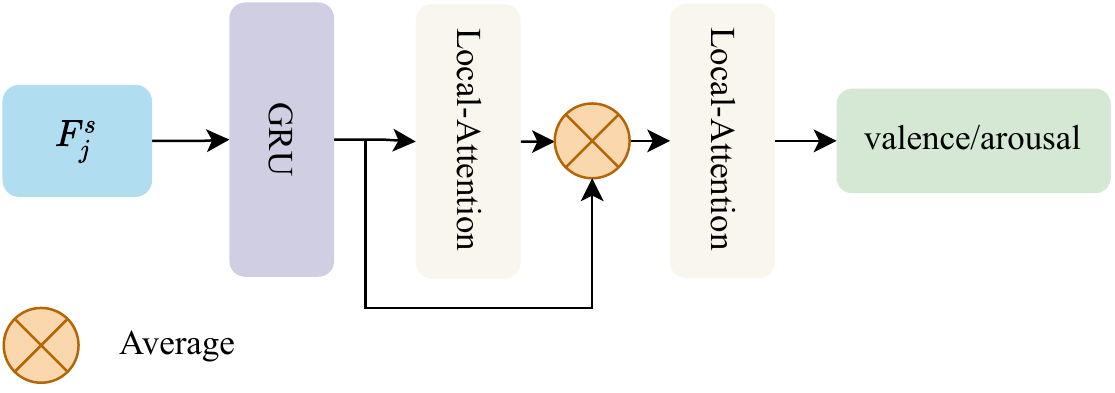}}
  \caption{Overview of prediction model: Gated Recurrent Unit combined with Local Attention. Where $s$ is sequence.}
  \label{fig:gru_att}
\end{figure}

We conducted K-fold cross validation to obtain different models for ensemble learning with $K=5$. The scores from each fold are combined together and to form a single vector for each frame in video, $F_{j}$, which can be formulated as
\begin{equation}
    F_j = \{V^1_{j,i},V^2_{j,i},V^3_{j,i},V^4_{j,i},V^5_{j,i},A^1_{j,i},A^2_{j,i},A^3_{j,i},A^4_{j,i},A^5_{j,i}\}
    \label{eqn:feat}
    \small
\end{equation}
where $V^{k}_{j,i}, A^{k}_{j,i}$ are valence and score for fold $k^{th}$ of $i^{th}$ frame in $j^{th}$ video with $k=\overline{1,K}$, $i=\overline{1,N_{j}}$, and $N_{j}$ is the length of $j^{th}$ video.
To ensemble results from K-fold models, we deployed GRU-based architecture for modelling temporal relationship, followed by two local attention layers to adjust the contribution of features, as in~\autoref{fig:gru_att}.

\subsection{Action unit detection}
\begin{figure*}
    \centering
    \includegraphics[width=.95\linewidth]{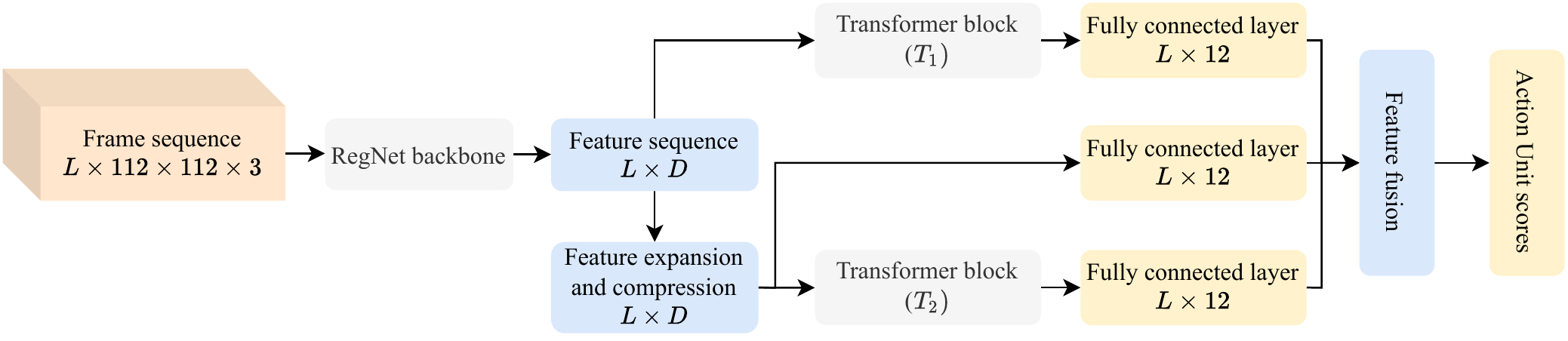}
    \caption{An overview of our action unit detection system.}
    \label{fig:au_sys}
\end{figure*}

In this task, two branches of feature are deployed with Transformer~\cite{vaswani2017attention} blocks, $T_{1}$ and $T_{2}$,~\autoref{fig:au_sys}. In $T_{2}$, the source feature are expanded to higher dimension and then compressed to original dimension with an aim to improve the robustness of the model. The new representation from this block also fed to a FC layer for creating an output to fuse with results from $T_{1}$ and $T_{2}$.

\section{Experiments and results}
\label{sec:experiments}
\subsection{Dataset}
The Valence-Arousal Estimate task includes 567 videos containing valence and arousal annotations. With 455 objects (277 males and 178 females) was annotated by four experts. The value range of valence and arousal is from -1 to 1. 

The Action Unit Detection task includes 548 videos annotating the six basic expressions, plus the neutral state, plus a category ’other’ that denotes expressions/affective states other than the six basic ones. Approximately 2,6 million frames, with 431 participants (265 males and 166 females), have been annotated by seven experts.
\subsection{Valence and arousal estimation}
GRU network was conducted with the PyTorch Deep Learning toolkit. The GRU was set with 256-dimensional hidden states, two-layer for multimodal and four-layered for prediction. The networks is trained in 25 epochs with initial learning rate of $0.001$ and Adam optimizer.
The mean between valence and arousal Concordance Correlation Coefficient (CCC), $\mathcal{P}$, is used to evaluate the performance of the model as
 \begin{equation}
    \begin{split}
        \label{eqn:va_combined}
        \mathcal{P}_{\mathrm{VA}} &= \frac{\mathcal{P}_\mathrm{V} + \mathcal{P}_{\mathrm{A}}}{2},
    \end{split}
\end{equation}
where $\mathcal{P}_\mathrm{V}$ and $\mathcal{P}_\mathrm{A}$ are the CCC of valence and arousal, respectively, which is defined as
\begin{equation}
    \begin{split}
        \label{eqn:ccc}
        \mathcal{P} &= \frac{2 \rho \sigma_{\hat Y} \sigma_{Y} }{\sigma_{\hat Y}^2 + \sigma_{\hat Y}^2 + (\mu_{\hat Y} - \mu_{Y})^2}
    \end{split}
\end{equation}

where  $\mu_{Y}$ was the mean of the label $Y$, $\mu_{\hat Y}$ was the mean of the prediction $\hat Y$, $\sigma_{\hat Y}$ and $\sigma_{Y}$ were the corresponding standard deviations,  $\rho$ was the Pearson correlation coefficient between $\hat Y$and $Y$. 

\begin{table}
  \centering
  \begin{tabular}{cccc}
    \toprule
    Fold & Valence & Arousal & Combined\\
    \midrule
    $1$ & 0.290 & 0.491 & 0.391\\
    $2$ & 0.348 & 0.435 & 0.391\\
    $3$ & 0.339 & 0.500 & 0.419\\
    $4$ & 0.294 & 0.491 & 0.392\\
    $5$ & 0.362 & 0.492 & 0.427\\
    Average & \textbf{0.390} & \textbf{0.540} & \textbf{0.465}\\
    Baseline~\cite{kollias2022abaw} & 0.31 & 0.17 & 0.24\\
    \bottomrule
  \end{tabular}
  \caption{Predicted results of Valence-Arousal, Combined ($0.5* Valence  + 0.5 * Arousal$). We used the RegNet feature and multimodal (GRU combined with Transformer) for training on the validation set.}
  \label{tab:kfold}
\end{table}

\begin{table*}[!htbp]
  \centering
  \begin{tabular}{ccccc}
    \toprule
    Method & Feature & $\mathcal{P}_{\mathrm{V}}$ & $\mathcal{P}_{\mathrm{A}}$ & $\mathcal{P}_{\mathrm{VA}}$ \\
    \midrule
    Baseline~\cite{kollias2022abaw} & ResNet & 0.31 & 0.17 & 0.24\\
    Deng et al.~\cite{deng2021iterative} & MobilefaceNet + MarbleNet & 0.442 & 0.546 & 0.494 \\
    Zhang et al.~\cite{zhang2021prior} & Expression Embedding~\cite{zhang2021learning} & \textbf{0.488} & 0.502 & 0.495 \\
    GRU + Transformer & RegNet & 0.391 & 0.565 & 0.478\\
    GRU & k-fold & 0.432 & 0.575 & 0.504 \\
    \textbf{GRU + Attention} & k-fold & 0.437 & \textbf{0.576} & \textbf{0.507}\\
    \bottomrule
  \end{tabular}
  \caption{The comparison with previous works on valence arousal estimation with Affwild2 validation set.}
  \label{tab:results}
\end{table*}
\autoref{tab:kfold} shown the results of our K-fold cross validation experiments with training set of Affwild2, and evaluate on Affwild2 validation set separately. All folds give better results than baseline with combined valence and arousal. In addition, the best result, $0.465$, is obtained by averaging 5 folds.

\autoref{tab:results} presented the results in the experiment. We conducted experiments both GRU separate and GRU combined with Local Attention by k-fold features. Moreover, we also experimented on GRU combined with Transformer by RegNet feature. All methods give better results than baseline by approximate twice time. The local attention made a small improvement of the ensemble model, which proving the potential of applying attention to the system. Furthermore, our method is higher than previous works ~\cite{deng2021iterative,zhang2021prior}, respectively 0.494 and 0.495. 

\subsection{Action Unit Detection}
Our network architectures is trained by using SGD with learning rate of $0.9$ combine and Cosine annealing warm restarts scheduler~\cite{loshchilov2016sgdr}. We optimized the network in 20 epochs with Focal loss function~\cite{lin2017focal} and evaluate with $F_{1}$ score.

\begin{table}[htbp]
    \centering
    \begin{tabular}{c c c} \toprule
        Method & Feature & $F_{1}$ \\ \midrule
        Baseline~\cite{kollias2022abaw} & VGG16 & 0.39 \\
        Our method & RegNet & 0.533 \\
        Our method with only $T_{1}$ & RegNet & 0.544 \\
        \bottomrule
    \end{tabular}
    \caption{The comprasion with prior methods on Affwild2 validation set for action unit detection.}
    \label{tab:my_label}
\end{table}
\section{Conclusions}
\label{sec:conclustion}
This paper utilized features from deep learning representations to the Valence-Arousal Estimation sub-challenge of ABAW3 2022. To extract information over time, GRU is used for sentiment analysis. To enhance the advantages of GRU, we have connected the Local Attention mechanism into our model. The CCC function was used to predict arousal/valence. Experimental results show that our proposed model outperforms the baseline method. We demonstrated that our results were better when combined GRUs and Local Attention. Furthermore, we introduced a new feature extracted from multimodal by combining folds. We showed that the new features improve not only speed but also accuracy. 

\section*{Acknowledgments}
This work was supported by the National Research Foundation of Korea (NRF)
grant funded by the Korea government (MSIT) (NRF-2020R1A4A1019191) and
Basic Science Research Program through the National Research Foundation of
Korea (NRF) funded by the Ministry of Education (NRF-2021R1I1A3A04036408).
{\small
\bibliographystyle{ieee_fullname}
\bibliography{egbib}

\begin{thebibliography}{10}\itemsep=-1pt

\bibitem{cho2014properties}
Kyunghyun Cho, Bart Van~Merri{\"e}nboer, Dzmitry Bahdanau, and Yoshua Bengio.
\newblock On the properties of neural machine translation: Encoder-decoder
  approaches.
\newblock {\em arXiv preprint arXiv:1409.1259}, 2014.

\bibitem{deng2020multitask}
Didan Deng, Zhaokang Chen, and Bertram~E Shi.
\newblock Multitask emotion recognition with incomplete labels.
\newblock In {\em 2020 15th IEEE International Conference on Automatic Face and
  Gesture Recognition (FG 2020)}, pages 592--599. IEEE, 2020.

\bibitem{deng2021iterative}
Didan Deng, Liang Wu, and Bertram~E Shi.
\newblock Iterative distillation for better uncertainty estimates in multitask
  emotion recognition.
\newblock In {\em Proceedings of the IEEE/CVF International Conference on
  Computer Vision}, pages 3557--3566, 2021.

\bibitem{deng2009imagenet}
Jia Deng, Wei Dong, Richard Socher, Li-Jia Li, Kai Li, and Li Fei-Fei.
\newblock Imagenet: A large-scale hierarchical image database.
\newblock In {\em 2009 IEEE conference on computer vision and pattern
  recognition}, pages 248--255. Ieee, 2009.

\bibitem{devlin2018bert}
Jacob Devlin, Ming-Wei Chang, Kenton Lee, and Kristina Toutanova.
\newblock Bert: Pre-training of deep bidirectional transformers for language
  understanding.
\newblock {\em arXiv preprint arXiv:1810.04805}, 2018.

\bibitem{kollias2022abaw}
Dimitrios Kollias.
\newblock Abaw: Valence-arousal estimation, expression recognition, action unit
  detection \& multi-task learning challenges.
\newblock {\em arXiv preprint arXiv:2202.10659}, 2022.

\bibitem{kollias2020analysing}
D Kollias, A Schulc, E Hajiyev, and S Zafeiriou.
\newblock Analysing affective behavior in the first abaw 2020 competition.
\newblock In {\em 2020 15th IEEE International Conference on Automatic Face and
  Gesture Recognition (FG 2020)(FG)}, pages 794--800.

\bibitem{kollias2019face}
Dimitrios Kollias, Viktoriia Sharmanska, and Stefanos Zafeiriou.
\newblock Face behavior a la carte: Expressions, affect and action units in a
  single network.
\newblock {\em arXiv preprint arXiv:1910.11111}, 2019.

\bibitem{kollias2021distribution}
Dimitrios Kollias, Viktoriia Sharmanska, and Stefanos Zafeiriou.
\newblock Distribution matching for heterogeneous multi-task learning: a
  large-scale face study.
\newblock {\em arXiv preprint arXiv:2105.03790}, 2021.

\bibitem{kollias2019deep}
Dimitrios Kollias, Panagiotis Tzirakis, Mihalis~A Nicolaou, Athanasios
  Papaioannou, Guoying Zhao, Bj{\"o}rn Schuller, Irene Kotsia, and Stefanos
  Zafeiriou.
\newblock Deep affect prediction in-the-wild: Aff-wild database and challenge,
  deep architectures, and beyond.
\newblock {\em International Journal of Computer Vision}, pages 1--23, 2019.

\bibitem{kollias2019expression}
Dimitrios Kollias and Stefanos Zafeiriou.
\newblock Expression, affect, action unit recognition: Aff-wild2, multi-task
  learning and arcface.
\newblock {\em arXiv preprint arXiv:1910.04855}, 2019.

\bibitem{kollias2021affect}
Dimitrios Kollias and Stefanos Zafeiriou.
\newblock Affect analysis in-the-wild: Valence-arousal, expressions, action
  units and a unified framework.
\newblock {\em arXiv preprint arXiv:2103.15792}, 2021.

\bibitem{kollias2021analysing}
Dimitrios Kollias and Stefanos Zafeiriou.
\newblock Analysing affective behavior in the second abaw2 competition.
\newblock In {\em Proceedings of the IEEE/CVF International Conference on
  Computer Vision}, pages 3652--3660, 2021.

\bibitem{kuhnke2020two}
Felix Kuhnke, Lars Rumberg, and J{\"o}rn Ostermann.
\newblock Two-stream aural-visual affect analysis in the wild.
\newblock In {\em 2020 15th IEEE International Conference on Automatic Face and
  Gesture Recognition (FG 2020)}, pages 600--605. IEEE, 2020.

\bibitem{lin2017focal}
Tsung-Yi Lin, Priya Goyal, Ross Girshick, Kaiming He, and Piotr Doll{\'a}r.
\newblock Focal loss for dense object detection.
\newblock In {\em Proceedings of the IEEE international conference on computer
  vision}, pages 2980--2988, 2017.

\bibitem{loshchilov2016sgdr}
Ilya Loshchilov and Frank Hutter.
\newblock Sgdr: Stochastic gradient descent with warm restarts.
\newblock {\em arXiv preprint arXiv:1608.03983}, 2016.

\bibitem{radosavovic2020designing}
Ilija Radosavovic, Raj~Prateek Kosaraju, Ross Girshick, Kaiming He, and Piotr
  Doll{\'a}r.
\newblock Designing network design spaces.
\newblock In {\em Proceedings of the IEEE/CVF Conference on Computer Vision and
  Pattern Recognition}, pages 10428--10436, 2020.

\bibitem{vaswani2017attention}
Ashish Vaswani, Noam Shazeer, Niki Parmar, Jakob Uszkoreit, Llion Jones,
  Aidan~N Gomez, {\L}ukasz Kaiser, and Illia Polosukhin.
\newblock Attention is all you need.
\newblock {\em Advances in neural information processing systems}, 30, 2017.

\bibitem{vu2021multitask}
Manh~Tu Vu, Marie Beurton-Aimar, and Serge Marchand.
\newblock Multitask multi-database emotion recognition.
\newblock In {\em Proceedings of the IEEE/CVF International Conference on
  Computer Vision}, pages 3637--3644, 2021.

\bibitem{zafeiriou2017aff}
Stefanos Zafeiriou, Dimitrios Kollias, Mihalis~A Nicolaou, Athanasios
  Papaioannou, Guoying Zhao, and Irene Kotsia.
\newblock Aff-wild: Valence and arousal ‘in-the-wild’challenge.
\newblock In {\em Computer Vision and Pattern Recognition Workshops (CVPRW),
  2017 IEEE Conference on}, pages 1980--1987. IEEE, 2017.

\bibitem{zhang2021prior}
Wei Zhang, Zunhu Guo, Keyu Chen, Lincheng Li, Zhimeng Zhang, and Yu Ding.
\newblock Prior aided streaming network for multi-task affective recognitionat
  the 2nd abaw2 competition.
\newblock {\em arXiv preprint arXiv:2107.03708}, 2021.

\bibitem{zhang2021learning}
Wei Zhang, Xianpeng Ji, Keyu Chen, Yu Ding, and Changjie Fan.
\newblock Learning a facial expression embedding disentangled from identity.
\newblock In {\em Proceedings of the IEEE/CVF Conference on Computer Vision and
  Pattern Recognition}, pages 6759--6768, 2021.

\end{thebibliography}
}

\end{document}